\documentclass[10pt,twocolumn,letterpaper]{article}

\usepackage{cvpr}
\usepackage{times}
\usepackage{epsfig}
\usepackage{graphicx}
\usepackage{amsmath}
\usepackage{amssymb}
\usepackage{subfigure}
\usepackage{multirow}
\usepackage{makecell}

\usepackage[pagebackref=true,breaklinks=true,letterpaper=true,colorlinks,bookmarks=false]{hyperref}

\cvprfinalcopy % *** Uncomment this line for the final submission

\def\cvprPaperID{2336} % *** Enter the CVPR Paper ID here

\ifcvprfinal\fi
\begin{document}

%%%%%%%%% TITLE
\title{Crowd Counting and Density Estimation by Trellis Encoder-Decoder Networks}

%\author{First Author\\
%Institution1\\
%Institution1 address\\
%{\tt\small firstauthor@i1.org}
%% For a paper whose authors are all at the same institution,
%% omit the following lines up until the closing ``}''.
%% Additional authors and addresses can be added with ``\and'',
%% just like the second author.
%% To save space, use either the email address or home page, not both
%\and
%Second Author\\
%Institution2\\
%First line of institution2 address\\
%{\tt\small secondauthor@i2.org}
%}
\author{Xiaolong Jiang\textsuperscript{1}\footnotemark[1] , Zehao Xiao\textsuperscript{1}\footnotemark[1] , Baochang Zhang\textsuperscript{3}, Xiantong Zhen\textsuperscript{5}, \\Xianbin Cao\textsuperscript{1,2}\footnotemark[2] , David Doermann\textsuperscript{4}, Ling Shao\textsuperscript{5}\\
\textsuperscript{1}School of Electronic and Information Engineering, Beihang University, Beijing, China, \\
\textsuperscript{2}Key Laboratory of Advanced technology of Near Space Information System (Beihang University), \\Ministry of Industry and Information Technology of China, Beijing, China \\
\textsuperscript{3}School of Automation Science and Electrical Engineering, Beihang University, Beijing, China, \\
\textsuperscript{4}Department of Computer Science and Engineering, University at Buffalo, New York, USA\\
\textsuperscript{5}Inception Institute of Artificial Intelligence, UAE\\
{\tt\small jasperj1tmac@163.com, zhxiao@buaa.edu.cn, bczhang@buaa.edu.cn, zhenxt@gmail.com,}\\
{\tt\small xbcao@buaa.edu.cn, doermann@buffalo.edu, ling.shao@ieee.org}}
% For a paper whose authors are all at the same institution,
% omit the following lines up until the closing ``}''.
% Additional authors and addresses can be added with ``\and'',
% just like the second author.
% To save space, use either the email address or home page, not both
%\and
%Second Author\\
%Institution2\\
%First line of institution2 address\\
%{\tt\small secondauthor@i2.org}

\maketitle
\thispagestyle{empty}

%%%%%%%%% ABSTRACT
\begin{abstract}
Crowd counting has recently attracted increasing interest in computer vision but remains a challenging problem. In this paper, we propose a trellis encoder-decoder network (TEDnet) for crowd counting, which focuses on generating high-quality density estimation maps. The major contributions are four-fold. First, we develop a new trellis architecture that incorporates multiple decoding paths to hierarchically aggregate features at different encoding stages, which improves the representative capability of convolutional features for large variations in objects. Second, we employ dense skip connections interleaved across paths to facilitate sufficient multi-scale feature fusions, which also helps TEDnet to absorb the supervision information. Third, we propose a new combinatorial loss to enforce similarities in local coherence and spatial correlation between maps. By distributedly imposing this combinatorial loss on intermediate outputs, TEDnet can improve the back-propagation process and alleviate the gradient vanishing problem. Finally, on four widely-used benchmarks, our TEDnet achieves the best overall performance in terms of both density map quality and counting accuracy, with an improvement up to $14\%$ in MAE metric. These results validate the effectiveness of TEDnet for crowd counting.%\footnote{The source code will be available after publication.}
\vspace{-4mm}
\end{abstract}
\renewcommand{\thefootnote}{\fnsymbol{footnote}}
\footnotetext[1]{These authors contribute equally.}
\footnotetext[2]{This author is the corresponding author.}

%%%%%%%%% BODY TEXT
\section{Introduction}
\label{sec:intro}
\begin{figure}[ht]
\centering
\includegraphics[scale=0.26]{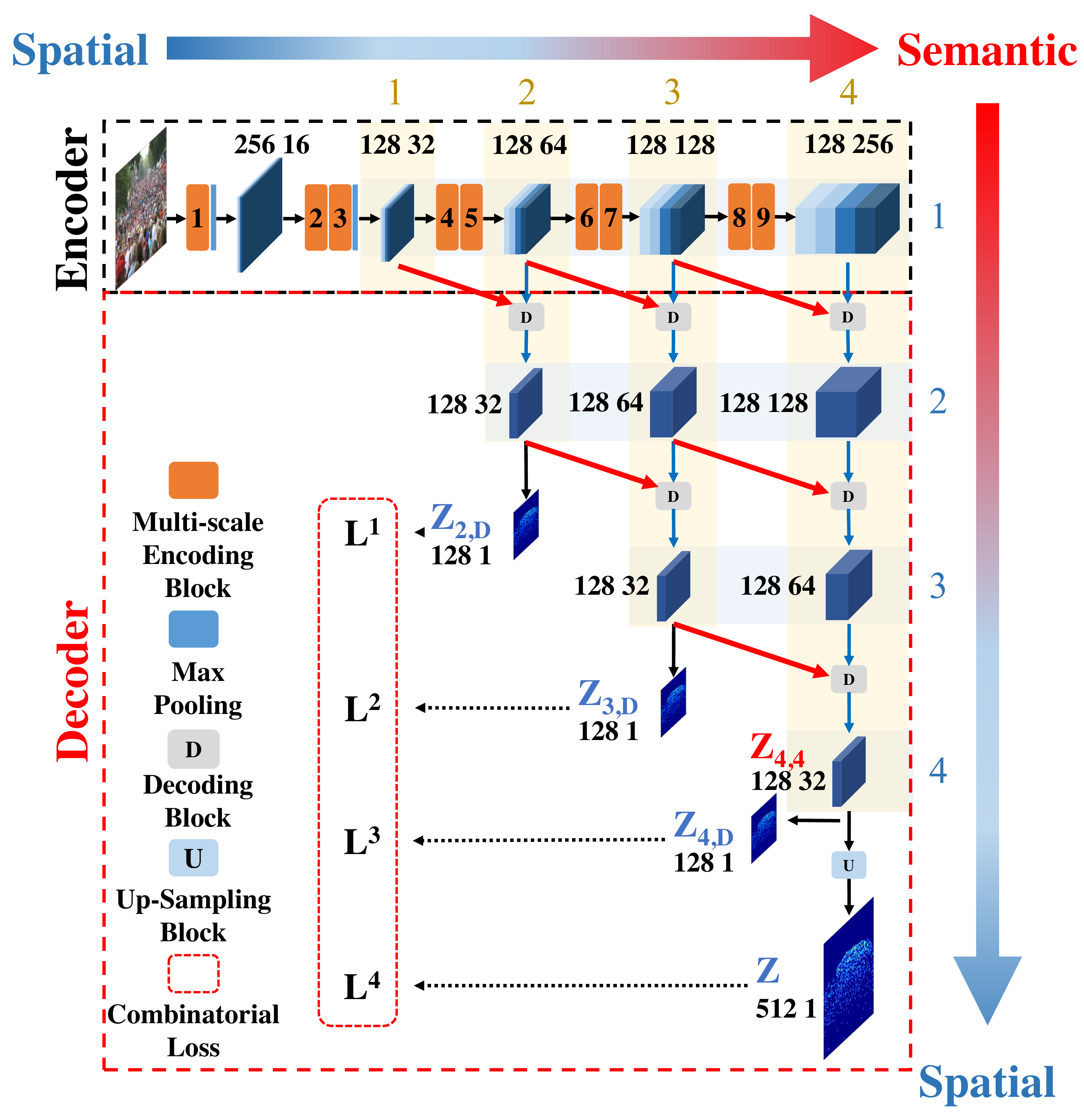}
\caption{An illustration of the Trellis Encoder-Decoder network (TEDnet) with distributed combinatorial losses. The horizontal and vertical axes indicate the spatial-semantic-spatial feature hierarchy established within TEDnet. We instantiate this hierarchy into a feature grid, whose rows and columns are indexed on the margin. The spatial and channel dimensions of each feature map is denoted by its side.}
\label{fig:Whole}
\vspace{-6mm}
\end{figure}
\begin{figure*}[ht]
\centering
\includegraphics[scale=0.43]{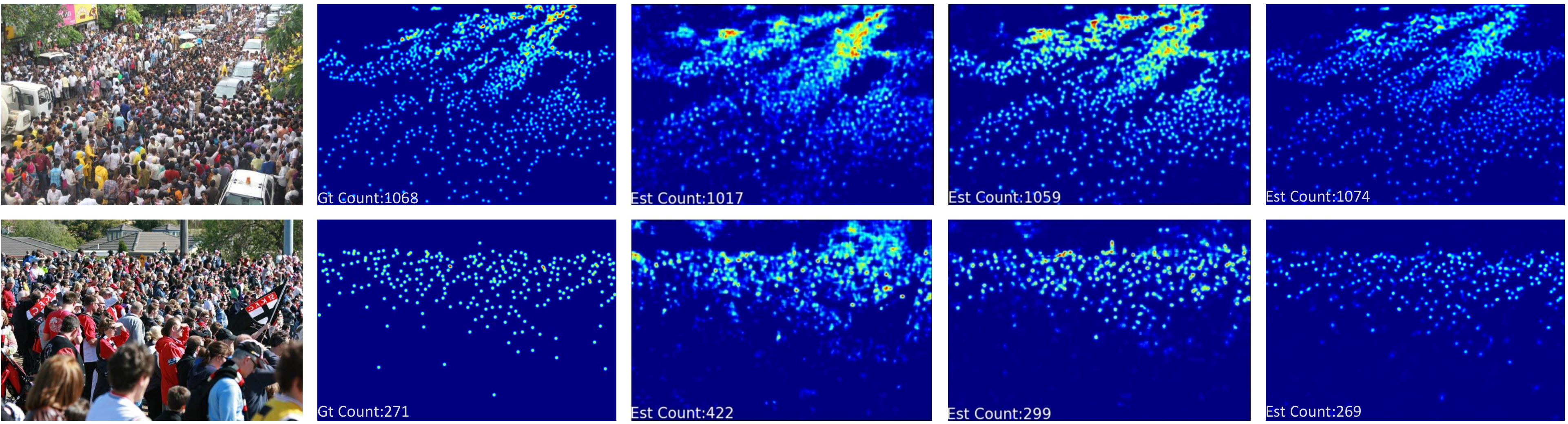}
\caption{An illustration of estimated density maps and crowd counts generated by the proposed approach and other state-of-the-arts. The first column shows two samples drawn from ShanghaiTech\_Part A. The second column shows corresponding ground truth maps with fixed Gaussian kernels. From the third to the last column we show the density maps estimated by MCNN \cite{MCNN}, SANet \cite{ECCV2018SANet}, and the proposed TEDNet, respectively. TEDnet generates density maps closer to the ground truth, and more accurate crowd counts.}
\label{fig:demo}
\vspace{-4mm}
\end{figure*}
With the rapid pace of urbanization, crowds tend to gather more frequently, increasing requirements for effective safety monitoring, disaster relief, urban planning, and crowd management. As a fundamental technique to support these applications, crowd counting has been investigated and has resulted in advanced solutions. Most crowd counting methods are based on detection \cite{2008DCC, 2009DCC, DCC2}, regression \cite{LinearRegressionCC, RidgeRegressionCC, Shah2013RCC}, and density estimation \cite{FirstDmapCC, MCNN, ECCV2018SANet, Arxiv2018CompositionLoss}. Density estimation based methods, in particular, have received increasing research focus. These techniques have the ability to localize the crowd by generating a density estimation map using pixel-wise regression. The crowd count is then calculated as the integral of the density map. To generate maps with a retained spatial size as the inputs, deep encoder-decoder convolutional neural network (CNN) architectures are widely applied \cite{SSFSRCNN, SS2, SegPSPnet, SegUnet, SDN, Refinenet, Refinenet2, wacvTrack}. In particular, encoder-decoder methods also play an important role in localization-oriented tasks to facilitate accurate pixel-wise regression \cite{CVPR2018CSRNet, DTCC, CFNet, cctracking, ccdetect2015cvpr}, given that the convolution itself is essentially a pixel-wise feature localization using traversal template matching. Thus motivated, we propose the trellis encoder-decoder network (TEDnet) for density estimation to address the crowd counting problem. Our approach improves both the encoding and decoding processes for more accurate pixel-wise estimations.

Feature encoding facilitates accurate pixel-wise estimations by extracting features, while preserving the pixel-wise localization precision in the feature maps. In scenes with severe occlusions and scale variations, CNN features are widely employed to enhance the feature encoding performance \cite{MCNN, Crowdnet, HydraCNN, PyramidCNNsICCV2017, ECCV2018SANet, Arxiv2018LeveragingUnlabeld}. It is worth noting that most current counting methods adopt CNNs that were originally designed for classification tasks, such as VGG-16 \cite{PyramidCNNsICCV2017, CVPR18DeepNegativeCorrelation, Arxiv2018LeveragingUnlabeld}, Inception \cite{ECCV2018SANet}, and DenseNet \cite{Arxiv2018CompositionLoss}. Despite their previous success, these networks build deep hierarchies to transform low-level spatial information into high-level semantic information. Consequentially, the resolution of feature maps is gradually degraded due to down-sampling operations, and thus, the localization precision is lowered. It is desirable to maintain a favorable balance between spatial resolution preservation and semantic feature extraction.

Feature decoding generates density maps by aggregating encoded feature maps. The pixel-wise accuracy performance for an estimated map is guaranteed by sufficient fusions of multi-scale decoding features that incorporate low-level spatial precision and high-level semantic depth. In hourglass encoder-decoder networks with a single decoding path \cite{SegUnet,SegPSPnet, SSFSRCNN, SS2}, features must endure excessive down-sampling and up-sampling operations, which degrade the pixel-wise precision. In addition, rich low-level spatial and high-level semantic information residing in multi-scale feature maps at the two ends of the hourglass are separated by the gap between the encoder and decoder. Although attempts have been made to enhance the hourglass networks with skip connections \cite{Refinenet, Refinenet2, SDN, SDN2}, they are not designed to generate high-quality density estimation maps due to the lack of hierarchical fusions between multi-scale features. From a more fundamental perspective, the widely adopted mean square error (MSE) loss in crowd counting assumes pixel-wise independence, while neglecting the local coherence and spatial correlation in density maps. It is therefore inadequate for facilitating the generation of high-quality density maps.

To address these issues in existing encoder-decoder networks and improve the counting performance with an enhanced architecture, we propose the trellis encoder-decoder network (TEDnet) to generate high-quality density maps. TEDnet achieves sufficient aggregation and fusion of multi-scale features within an established trellis-like feature hierarchy. In the encoding process, multi-scale convolutional kernels are used to obtain scale adaptation, where down-sampling strides are cut to four to preserve pixel-wise spatial precision. In the decoding process, multiple paths are deployed at corresponding encoding stages, each of which aggregates the encoded multi-scale features. Across paths, features containing diverse spatial and semantic information are integrated using dense skip connections, which guarantees thorough multi-scale feature fusions. Our multi-path trellis network is similar in spirit to an ensemble of multiple hourglass networks with different feature scales, establishing a feature learning hierarchy resides in a trellis structure, as highlighted in Figure \ref{fig:Whole}. Each path in TEDnet generates an intermediate output map that intrinsically enables the deployment of distributed supervision within each path. This alleviates the gradient vanishing problem and boosts the gradient flow through the network. Each distributed loss in TEDnet is a combinatorial loss defined based on the proposed spatial abstraction loss (SAL) and the spatial correlation loss (SCL). SAL and SCL relieve the pixel-wise independence assumption posed by the MSE loss, and improve the density map quality, as well as counting performance, by enforcing similarities in local coherence and spatial correlation between maps.

TEDnet takes full images, rather than image patches, as the input and outputs full-resolution density maps. This further ensures the density map quality (qualitatively demonstrated in Figure~\ref{fig:demo}) by avoiding the tedious patch-wise operation, which induces boundary artifacts. The main contributions of the proposed approach are summarized as follows:
\begin{itemize}
\item We propose a new deep learning architecture for accurate density estimation and crowd counting, called trellis encoder-decoder network (TEDnet), which assembles multiple encoding-decoding paths hierarchically to generate high-quality density map for accurate crowd counting.
\item We establish a multi-path decoder that pervasively aggregates the spatially-endowed features within a decoding feature hierarchy and progressively fuses multi-scale features with dense skip connections interleaved in the hierarchy.
\item We introduce a combinatorial loss comprising of the newly designed SAL and SCL to supervise local coherence and spatial correlation in density maps. Distributed supervision, in conjunction with the combinatorial loss, is deployed on intermediate multi-path outputs to improve the optimization of the network.
\item We achieve the best overall performance on four commonly-used benchmark datasets, largely surpassing the state-of-the-art methods by up to $14\%$ for the MAE metric. We obtain the best quality of estimated density maps, in terms of both PSNR and SSIM measures.
\end{itemize}

%\vspace{1mm}
\section{Related Work}
\label{sec:related}
\vspace{-1mm}
In this section, we provide a brief review of the most related work and refer to comprehensive surveys for crowd counting \cite{Survey2015a, Survey2015b, Survey2018, beyondcounting}.

\vspace{-1mm}
\subsection{Detection and Regression based Methods}
\vspace{-1mm}
Detection-based counting methods deploy a detector to traverse the image, which localizes and counts the targets along the way \cite{2012HumanDetection, 2009DCC, 2008DCC, DCC2}. These methods are surpassed by the regression-based alternatives, as the detection performance is affected in the presence of over-crowded scenes. The successes of regression-based methods \cite{RCC1, RCC2, Shah2013RCC, ZhangCNNPatch, MoCNN} can thus be attributed to their ability of circumventing explicit detection and directly mapping the input images to scalar values. Nevertheless, regression-based methods forfeit localization capability such that they cannot perceive crowd distributions. To recover the lost localization capability, crowd counting methods based on density estimation are therefore developed by conducting pixel-wise regressions.

\vspace{-1mm}
\subsection{Density Estimation based Methods}
\vspace{-1mm}
Initially introduced in \cite{FirstDmapCC}, density estimation based methods avoid explicitly detecting each individual and retain the ability to localize the crowd. Earlier approaches strove to compute density maps with hand-crafted features \cite{FirstDmapCC, DensityCCNoDL2} and random forest regressions \cite{DensityCCNoDL2, DensityCCNoDL3, DensityCCNoDL4}. More recent methods appeal to CNN based feature extraction to supply scale and perspective invariant features. In particular, MCNN \cite{MCNN}, Crowdnet \cite{Crowdnet}, Hydra CNN \cite{HydraCNN}, CNN-boost \cite{CNNboost}, CP-CNN\cite{PyramidCNNsICCV2017}, and Switching CNN \cite{SwitchingCNN} all conform to an ensemble design approach to enable multi-scale adaptation, where multiple CNN branches with different receptive fields are jointly maintained. The extra computational expense introduced by these methods is to some degree wasted on inefficient and un-flexible branching \cite{CVPR2018CSRNet}. As a remedy, single-branch counting networks with scale adaptations were proposed in \cite{ECCV2018SANet, Arxiv2018CompositionLoss, ourcc}. Notably, most of these methods follow a patch-based counting mechanism \cite{CNNpatch, CNNboost, HydraCNN, Crowdnet, SwitchingCNN, Decidenet, Arxiv2018CompositionLoss, ECCV2018SANet}, where the full density map is obtained by concatenating discrete density patches. More importantly, methods such as MCNN, Hydra CNN, and CNN-boost output density maps with reduced resolution due to excessive down-sampling strides. This inevitably sacrifices pixel-wise details and damages the density map quality. Comparatively, CP-CNN \cite{PyramidCNNsICCV2017} focuses on generating high-quality full-resolution maps with the help of global and local semantic information. In \cite{ECCV2018SANet}, researchers computed high-quality full-resolution maps with a new encoder-decoder network, as well as a SSIM local pattern consistent loss. In order to limit the down-sampling stride in the encoding process, CSRNet \cite{CVPR2018CSRNet} adopts dilated convolutional layers to substitute pooling layers.

Unlike other approaches, the proposed trellis encoder-decoder architecture attempts to generate high-quality density estimation maps by preserving the spatial information in the encoding feature hierarchy. More importantly, it incorporates a multi-path decoder to enhance the aggregation and fusion of multi-scale features with rich spatial and semantic information. As a result, pixel-wise regression accuracy in the estimated map is enhanced. In a broader view, density estimation is similar to other localization-oriented tasks, such as tracking \cite{cctracking2011iccv, cctracking, LaoxieACCV} and detection \cite{ccdetect2015cvpr}, which also generate localization estimation maps as outputs. These tasks are inter-correlated with density estimations such that the resulting localization maps can be fused to integrate task-specific localization response \cite{beyondcounting, Arxiv2018CompositionLoss}. Moreover, semantic segmentation also relies on powerful encoder-decoder architecture to integrate multi-scale features and to improve localization precision. Consequently, efforts have been made to enhance the hourglass architecture. In \cite{SDN}, SDN stacks multiple single-path hourglass networks into a deeper sequence to improve the feature fusion and guarantee fine recovery of localization information. In \cite{Refinenet, Refinenet2}, the single-path hourglass network is extended by adding residual units inside the skip connections.
\vspace{-2mm}
\section{Trellis Encoder-Decoder Networks}
\vspace{-2mm}
As shown in Figure \ref{fig:Whole}, the goal of TEDnet is to achieve improved counting performance by generating density maps with high pixel-wise density estimations. In the encoder, the localization property of a density estimation conforms to the nature of a convolutional layer operation. Here, the convolutional kernels are the feature templates that are localized in the feature maps via template-matching. In the decoder, encoded feature maps are aggregated to represent the locality of crowded objects. Our TEDnet can establish a feature hierarchy within the trellis architecture, where reliable multi-scale features are encoded with well-preserved spatial information. These are then decoded into accurate density maps, with a great capacity for precise localization. In what follows, we explain in detail the multi-scale encoder, the multi-path decoder, and the distributed supervision with combinatorial loss in TEDnet.

\vspace{-2mm}
\subsection{Multi-Scale Encoder}
\vspace{-2mm}
We design the multi-scale encoder to extract reliable features relevant to crowded human objects, while being able to localize these features with pixel-wise precision. The multi-scale encoding block is capable of overcoming occlusions and scale variations present in crowd counting scenes, as elaborated below.

\begin{figure}[ht]
\centering
\includegraphics[scale=0.16]{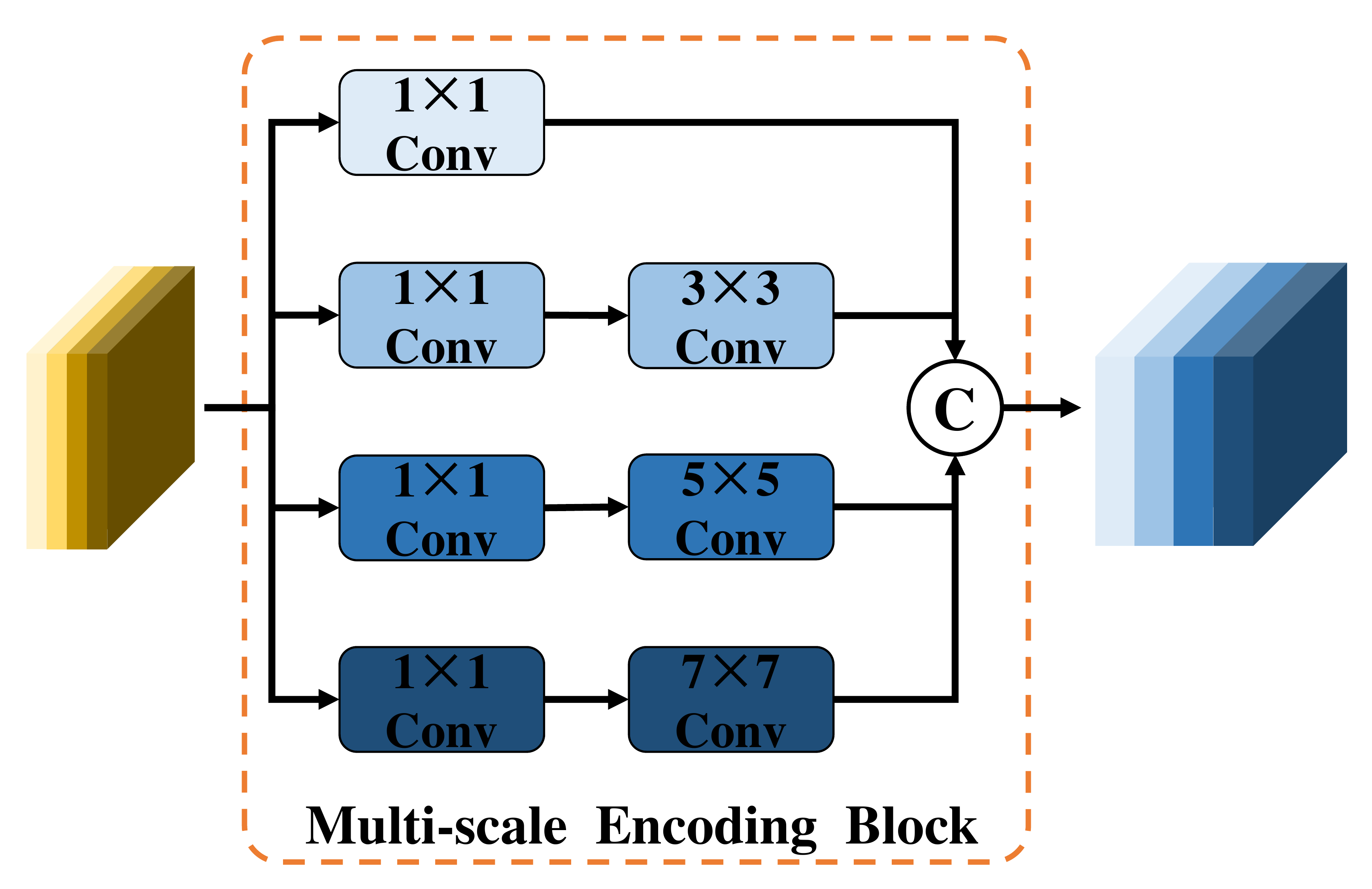}
\caption{An illustration of the multi-scale encoding block. Circled C in the figure represents channel-wise concatenation.}
\label{fig:EncodingBlock}
\vspace{-7mm}
\end{figure}

As shown in Figure~\ref{fig:EncodingBlock}, a multi-scale encoding block is implemented with kernels of different sizes, which enables the encoder to extract multi-scale features. As indicated in Figure~\ref{fig:Whole}, a total of nine encoding blocks are implemented and grouped into five encoding stages. To preserve feature localization precision, we limit the application of pooling operations. Consequently, only two $2 \times 2$ max pooling layers are inserted at the first two encoding stages, each of which has a down-sampling stride of $2$. To further enlarge the receptive fields, dilated convolutional kernels with dilation rates of $2$ and $4$ are employed in the last two encoding blocks \cite{CVPR2018CSRNet}.

\subsection{Multi-Path Decoder}

We design a new multi-path decoder to hierarchically aggregate the spatially-preserved features and restore the spatial resolution in the density map. As the component that directly generates the density maps, the decoder has a vital influence on the density map quality. Unfortunately, less emphasis had been placed on the decoder in the literature for crowd counting and density estimation. In CSRNet \cite{CVPR2018CSRNet}, the density maps are simply generated by applying bilinear interpolation to up-sample the encoded feature maps.

In Crowdnet \cite{Crowdnet}, a $1 \times 1$ convolutional layer is implemented as the decoder. SANet \cite{ECCV2018SANet} advocates the idea of refinement in a single-path hourglass decoder. To the best of our knowledge, it is by far the most sophisticated design in the context of density estimation. Alternatively, efforts have been made in other tasks using the hourglass architecture, such as image segmentation \cite{SegUnet,SegPSPnet, SDN, Refinenet} and super-resolution \cite{SSFSRCNN, SS2}. Nevertheless, as explained in Section \ref{sec:intro}, these architectures are not optimal for density estimation. They suffer from prolonged single-path feature transformation hierarchy with heavy parameterization, as well as insufficient feature aggregations and fusions.

To remedy the defects of the existing decoder, we propose a multi-path decoder in TEDnet, which assembles a set of single-path hourglass architectures with multi-scale features. As depicted in Figure~\ref{fig:Whole}, three decoding paths are exploited on the feature maps, computed from the last three encoding stages. Within each path, a decoding feature hierarchy is established to aggregate feature representations at the same semantic level, in a progressive way. Among different paths, feature maps from different levels are fused with dense skip connections. Both aggregation and fusion for features are implemented in densely interleaved decoding blocks. The decoder implementation is realized by stacking decoding blocks into the trellis structure, such that a feature hierarchy is established. As shown in Figure~\ref{fig:Whole}, such a feature hierarchy is pinpointed into the trellis architecture with a grid representation, where each column indicates one decoding path and each row presents the depth within each path.

\vspace{-5mm}
\begin{figure}[htbp]
\centering
\subfigure[]{
\label{fig:Decoding}
\begin{minipage}{3cm}%[t]
\centering
\includegraphics[scale=0.15]{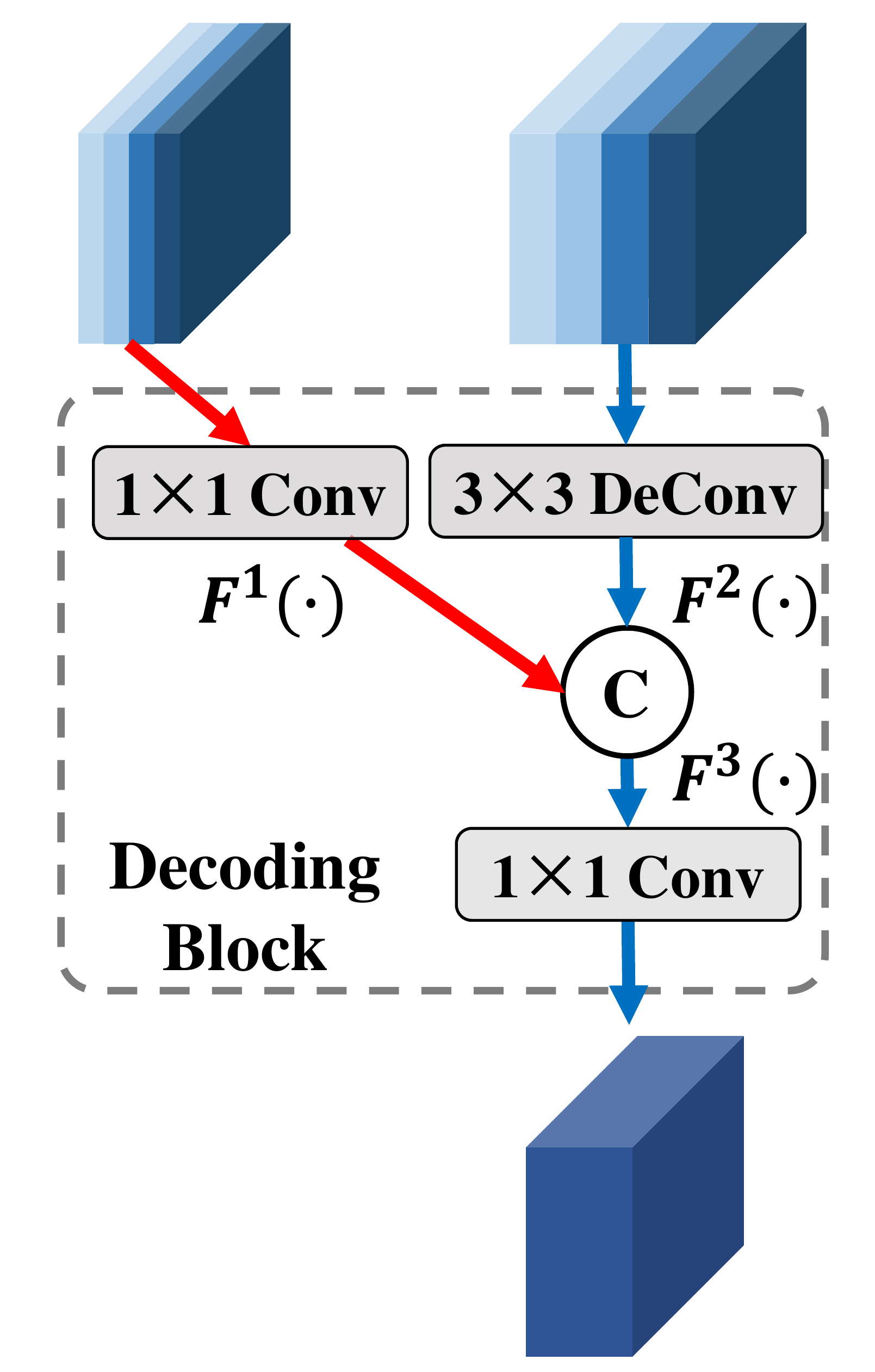}
%\caption{The illustration of the decoding block. Circled C in the figure represents channel-wise concatenation.}
%\label{fig:Decoding}
%\vspace{-4mm}
\end{minipage}
}
\subfigure[]{
\label{fig:UpsampleBlock}
\begin{minipage}{3cm}%[t]
\centering
\includegraphics[scale=0.157]{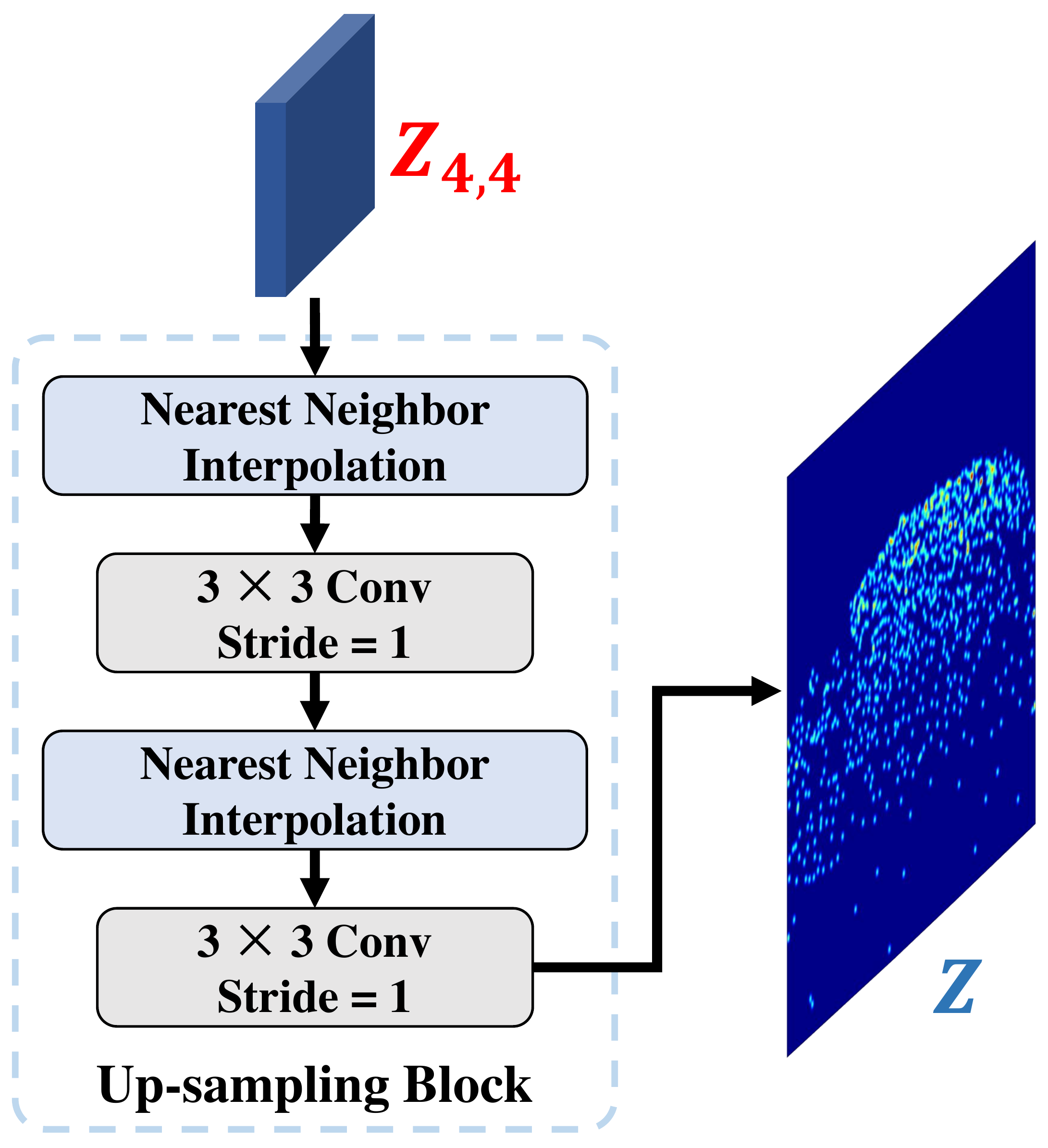}
\end{minipage}
}
\caption{An illustration of the decoding block (a) and the up-sampling block (b). Circled C in the figure represents channel-wise concatenation.}
\vspace{-5mm}
\end{figure}
\paragraph{Decoding Block.} As shown in Figure~\ref{fig:Decoding}, each decoding block takes two inputs. The right input feature is passed from the same decoding path and it possesses deeper semantic information whose channels doubles those of the left input feature. It is aggregated via a deconvolutional layer $F_{i,j}^2$ with $3 \times 3$ kernels, which halves the channels. The left input feature is aggregated by a convolutional layer $F_{i,j}^1$ deploying $1 \times 1$ kernels, with its depth unchanged. These two aggregated features are fused through channel-wise concatenation, followed by a convolutional layer $F_{i,j}^3$ with $1 \times 1$ filters.

In (\ref{equa:decoder}), $Z_{i,j}$ denotes the decoded feature at the $i$-th row and $j$-th column within the feature grid, computed by the decoding block $D_{i,j}$, as follows:
\begin{equation}
\begin{aligned}
{Z_{{\rm{i}},j}} &= {D_{i,j}}({Z_{i - 1,j - 1}},{Z_{i - 1,j}}) \\
&= F_{i,j}^3([F_{i,j}^1({Z_{i - 1,j - 1}}),F_{i,j}^2({Z_{i - 1,j}})]),\\
%&i,j \in \{ 2,3,4\}
\end{aligned}
\label{equa:decoder}
\end{equation}
where $F(\cdot)$ indicates a convolutional operations, and $[\cdot]$ denotes a channel-wise concatenation.

Within the established feature hierarchy as shown in Figure~\ref{fig:Whole}, the decoded features enable the aggregation and fusion of multi-scale features. As a result, the decoded feature map $Z_{4,4}$, at the end of the rightmost decoding path, contains the richest spatial and semantic information. Thus, the final output density map $Z$ is generated from these feature maps by restoring the spatial dimension through the up-sampling block.
\vspace{-5mm}
\paragraph{Up-sampling Block.} As illustrated in Figure \ref{fig:UpsampleBlock}, the design of the up-sampling block is inspired by a super-resolution technique \cite{SISRupample}, where the nearest neighbor interpolation is followed by a $3\times3$ convolutional layer with a stride of $1$. The overall down-sampling stride of TEDnet is $4$. We restore the spatial size of the density map by repeating the up-sampling operations twice in the up-sampling block.

Overall, a spatial-semantic-spatial feature hierarchy is fully exploited in TEDnet. In Figure~\ref{fig:Whole}, the proposed architecture is established to host the feature hierarchy. As indicated by the horizontal axis, the feature maps on the right in the hierarchy have more semantic information than the ones on the left. Those on the left, however, contain richer spatial details. Vertically, spatial information is gradually recovered through skip connections, which transmit low-level spatial features from left to right, top to bottom. It is worth noting that, for a simple single-path hourglass encoder-decoder, spatial information cannot be recovered in the decoder as indicated vertically in Figure~\ref{fig:Whole}. Although sparsely linked skip connections can alleviate inadequate feature fusion to a certain extent in single-path hourglass encoder-decoders, the pervasive feature fusions as realized in TEDnet can still not be reached.
\vspace{-2mm}
\subsection{Distributed Supervision}
\begin{figure}[t]
\centering
\includegraphics[scale=0.14]{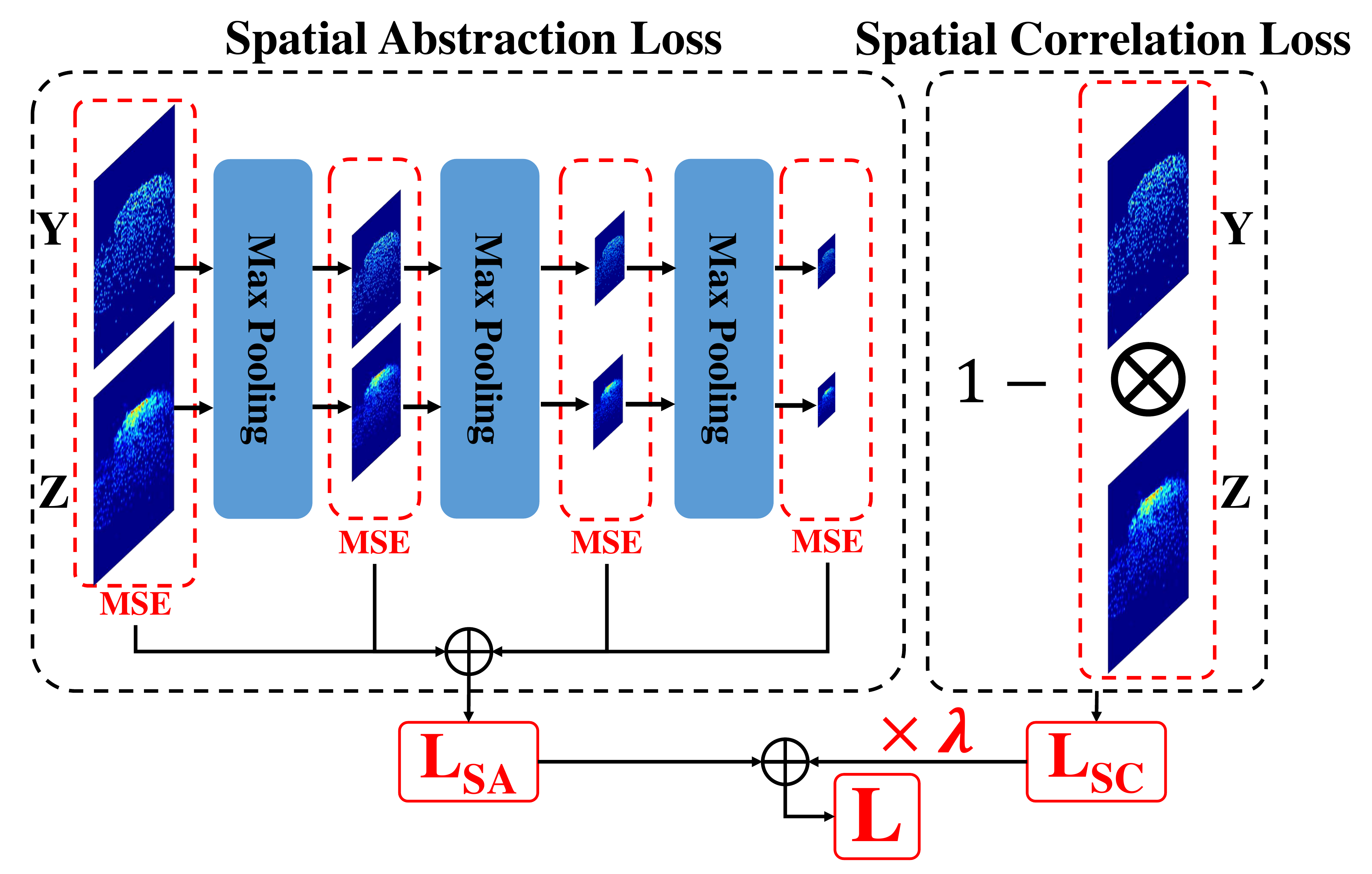}
\caption{An illustration of the combinatorial loss $L$. $Z$ indicates one of the distributed outputs, while $Y$ is the corresponding ground truth map, resized to be the same as $Z$.}
\label{fig:Loss}
\vspace{-6mm}
\end{figure}
\vspace{-2mm}
The multi-path architecture of TEDNet produces intermediate output estimation maps, i.e., $Z_{2,D}$, $Z_{3,D}$, $Z_{4,D}$, $Z$, at the ends of decoding paths as illustrated in Figure  \ref{fig:Whole}.
This design naturally enables distributed supervision, such that multiple losses can be applied at each intermediate output. Previous attempts have been made to provide multi-supervision, where losses are computed between intermediate feature maps and the ground truths \cite{MultiSup1hed, MultiSup2a, SDN}. In contrast, the proposed distributed supervision implemented in TEDnet computes multiple losses between intermediate density estimation maps and ground truth maps. From the ensemble point of view, each distributed loss is calculated to supervise the corresponding path representing a single-path hourglass network. In particular, to compute the losses at $Z_{2,D}$, $Z_{3,D}$, $Z_{4,D}$, each of them is aggregated from its previous feature map using a convolutional layer with $1 \times 1$ filter size. The ground truth density map is down-sampled to $128 \times 128$ with average pooling operations. Each of these intermediate outputs is separately decoded on different feature levels along its own path. Meanwhile, information from different paths are integrated through dense skip connections. As a result, the supervision at each output is meaningful and can help better optimize the network.

Due to the distributed supervision, in conjunction with the dense skip connections, the gradient vanishing phenomenon, which indicates weaker gradients at the earlier stage of the network, is substantially alleviated. Consider the convolutional block $1$ for instance. During the back-propagation process, the gradients flow is a summation of propagated flows, starting at each distributed supervision, such that the gradient is boosted. Moreover, for each flow originating at its corresponding supervision, instead of flowing backward along just one decoding path, the interleaved dense skip connections provide more diffused flow paths at each fork junctions, thus further boosting the gradient flow.
\begin{figure*}[t]
\centering
\includegraphics[scale=0.30]{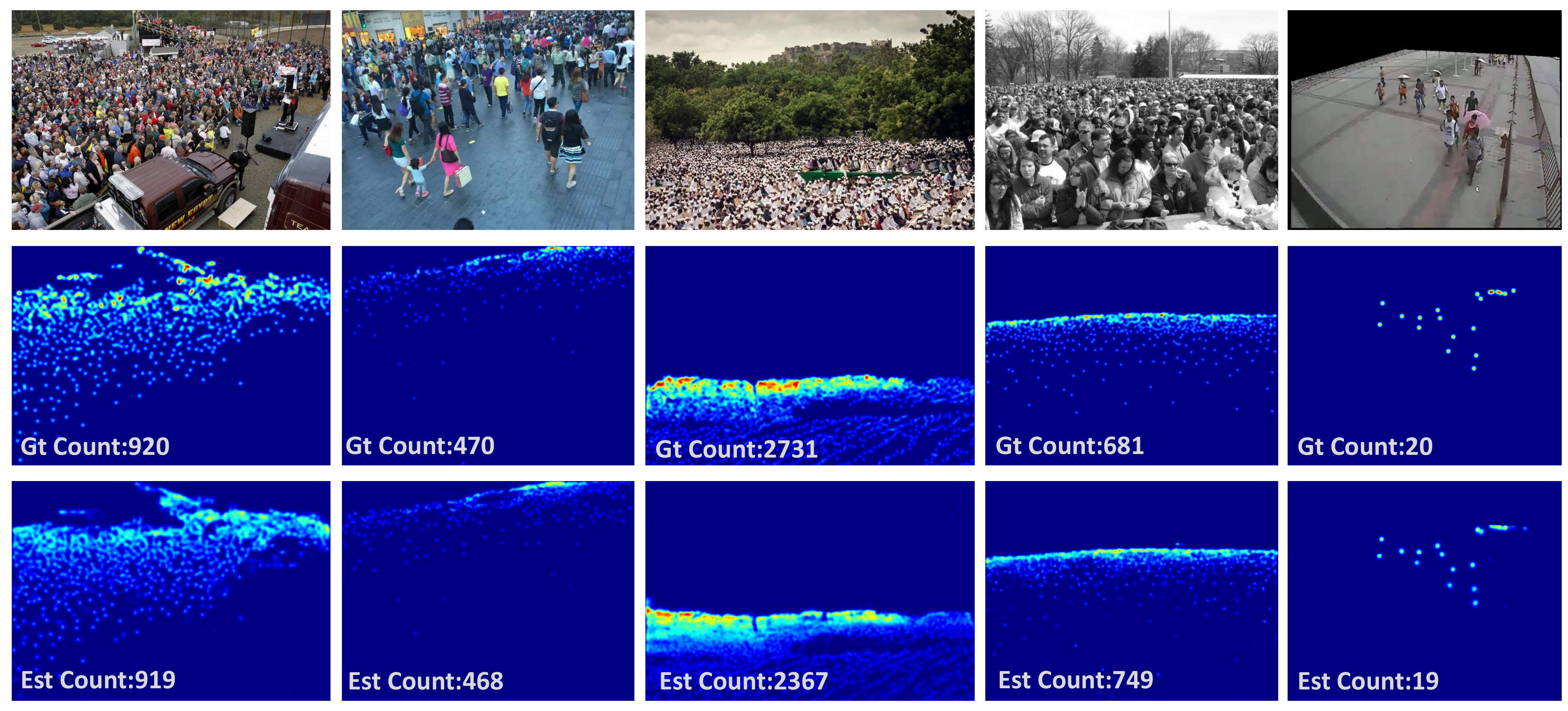}
\caption{From left to right, we display the density maps generated by TEDnet on ShanghaiTech Part\underline{ }A, ShanghaiTech Part\underline{ }B, UCF-QNRF, UCF\underline{ }CC\underline{ }50, and WorldExpo'10 datasets. The second row shows ground truth density map, the third row depicts our estimated maps.}
\label{fig:LastDemo}
\vspace{-6mm}
\end{figure*}
\vspace{-6mm}
\subsection{Combinatorial Loss}
\vspace{-2mm}
As shown in Figure~\ref{fig:Loss}, the loss function distributed at each decoding output is a combination of two losses. In general, the pixel-wise mean square error (MSE) loss has dominated the training of density estimation based crowd counting approaches \cite{MCNN, Crowdnet, Arxiv2018CompositionLoss}. As advocated in \cite{ECCV2018SANet, CVPR2018CSRNet}, the MSE loss assumes pixel-wise isolation and independence. As a result, it is incapable of enforcing spatial correlation and coherence among pixels in the estimated maps, which, however, plays an important role in influencing the quality of the density map. To compensate the limited MSE loss, we define a spatial abstraction loss (SAL) and a spatial correlation loss (SCL), resulting in a combinatorial loss.
\vspace{-6mm}
\paragraph{Spatial Abstraction Loss.} SAL progressively computes the MSE losses on multiple abstraction levels between the predicted map and the ground truth. These spatial abstractions are instantiated by cascading max pooling layers with down-sampling strides, leading to a gradually enlarged receptive field on each level. At each level, the pixel value in an abstracted map is non-linearly sampled from a receptive field at the corresponding location in the preceding abstraction level. By computing MSE on each abstraction level, SAL can supplement the pixel-wise MSE loss with patch-wise supervision. In our experiments, after a normal MSE loss, we implement three levels of abstraction ($K = 3$ in the following equation) with $2 \times 2$ max pooling layers, each with a stride of $2$. The computation of SAL is formalized as:
\begin{equation}
{{L}_{SA}}=\sum\limits_{k=1}^{K}\frac{1}{N_{k}}{\left\| {{\varphi }_{k}}(Z)-{{\varphi }_{k}}(Y) \right\|}_{2}^{2},
\end{equation}
where $\varphi_{k}(\cdot)$ denotes the abstraction computation on the $k$-th abstraction level. $N_{k}$ is the number of pixels within a map on the $k$-th abstraction level.
%\vspace{2mm}
\vspace{-6mm}
\paragraph{Spatial Correlation Loss.} Beyond the patch-wise supervision enforced by SAL, SCL further complements the pixel-wise MSE loss with map-wise computation. SCL represents the difference between two density maps based on normalized cross correlation (NCC) similarity. This is less sensitive to linear changes in the density map intensity. In addition, SCL is easier to compute and experimentally friendly compared to the MSE loss. The computation of SCL defined on two maps is:
\begin{equation}
%{L_{SC}} = 1 - \frac{{\sum\limits_i^I {\sum\limits_j^J {({Z_{ij}} \cdot {Y_{ij}})} } }}{{\sqrt {\sum\limits_i^I {\sum\limits_j^J {Z_{ij}^2} }  \cdot \sum\limits_i^I {\sum\limits_j^J {Y_{ij}^2} } } }} ,
{L_{SC}} = 1 - \frac{{\sum\limits_p^P {\sum\limits_q^Q {({Z_{pq}} \cdot {Y_{pq}})} } }}{{\sqrt {\sum\limits_p^P {\sum\limits_q^Q {{Z_{pq}}^2} }  \cdot \sum\limits_p^P {\sum\limits_q^Q {{Y_{pq}}^2} } } }} ,
\end{equation}
where $Y_{pq}$ and $Z_{pq}$ represent the pixels in the ground truth density map and the predicted density map, respectively. $p$ and $q$ are the row and column indexes in the map, and $P \times Q$ denotes the total number of pixels.
The final combinatorial loss $L$ is formulated as a weighted sum of SAL and SCL as:
\begin{equation}
{{L}}={{L}_{SA}}+\lambda {{L}_{SC}} ,
\end{equation}
where $\lambda$ is a factor to balance the contributions of SAL and SCL. The selection of $\lambda$ is explained in Section \ref{sec:42}.
\vspace{-2mm}
%-------------------------------------------------------------------------
\section{Experiments and Results}
%In this section, we first provide implementation details as well as the performance metrics deployed in our experiments. We then present an ablation study to demonstrate the performance of each contribution of TEDnet, and a comprehensive comparison with the state-of-the-arts is conducted. Arrows in the tables indicate the favorable directions of the metric values.
\vspace{-1mm}

\subsection{Implementation details}
\vspace{-2mm}
Following \cite{ourcc}, we generate our ground truth maps by fixed size Gaussian kernels and augment the training data with an online sampling strategy (more details can be found in \cite{ourcc}). We train our TEDnet in an end-to-end manner from scratch, and optimize the network parameters based on the Adam optimizer \cite{GTGen}. We use batch size of 8, Xavier initialization, and an initial learning rate of $1e-3$. The learning rate is step-wise and decreased by a factor of $0.8$ every 10K iterations. In regard to the efficiency performance of TEDnet, it takes 2500 epochs for training to convergence, and 0.027s for testing each image on ShanghaiTech Part\underline{ }A.

\vspace{-4mm}
\paragraph{Image-wise Operation.}To generate high-quality full-resolution density maps, TEDnet takes full-size images as inputs and outputs the same size density maps. Our approach differs from methods that adopt patch-wise operations \cite{CNNpatch, CNNboost, HydraCNN, Crowdnet, SwitchingCNN, Decidenet, Arxiv2018CompositionLoss, ECCV2018SANet}. Notably, patch-wise operations induce boundary artifacts that negatively affect the localization precision. Moreover, the patch-wise counting accuracy suffers from statistical shifts across patches \cite{ECCV2018SANet}.
\vspace{-5mm}
\paragraph{Counting Accuracy.}To evaluate the counting accuracy, we adopt the mean average error (MAE) and the mean squared error (MSE) metrics, which are defined as:
\begin{equation}
%\small
\footnotesize
{\mathrm{MAE} = \frac{1}{M}\sum\limits_{i = 1}^M {|{C_i} - C_i^{gt}|}}, ~~~  \mathrm{MSE} = \sqrt{\frac{1}{M}\sum\limits_{i = 1}^M {|{C_i} - C_i^{gt}{|^2}}}
\end{equation}
where $M$ is the number of images in the test set, and $C_i^{gt}$ and $C_i$ represent the ground truth and the predicted count of the $i$-th image, computed as the integral of the density maps.
\vspace{-3mm}
\paragraph{Density Map Quality.}To evaluate the quality of the estimated density maps, we also calculate the PSNR (Peak Signal-to-Noise Ratio) and SSIM (Structural Similarity in Image) indices, as described in \cite{PyramidCNNsICCV2017}. In particular, the SSIM index is normally adopted in image quality assessment \cite{ssim}, and it computes the similarity between two images from the mean, variance and covariance statistics.
\vspace{-2mm}
\begin{figure}[t]
\centering
\includegraphics[scale=0.36]{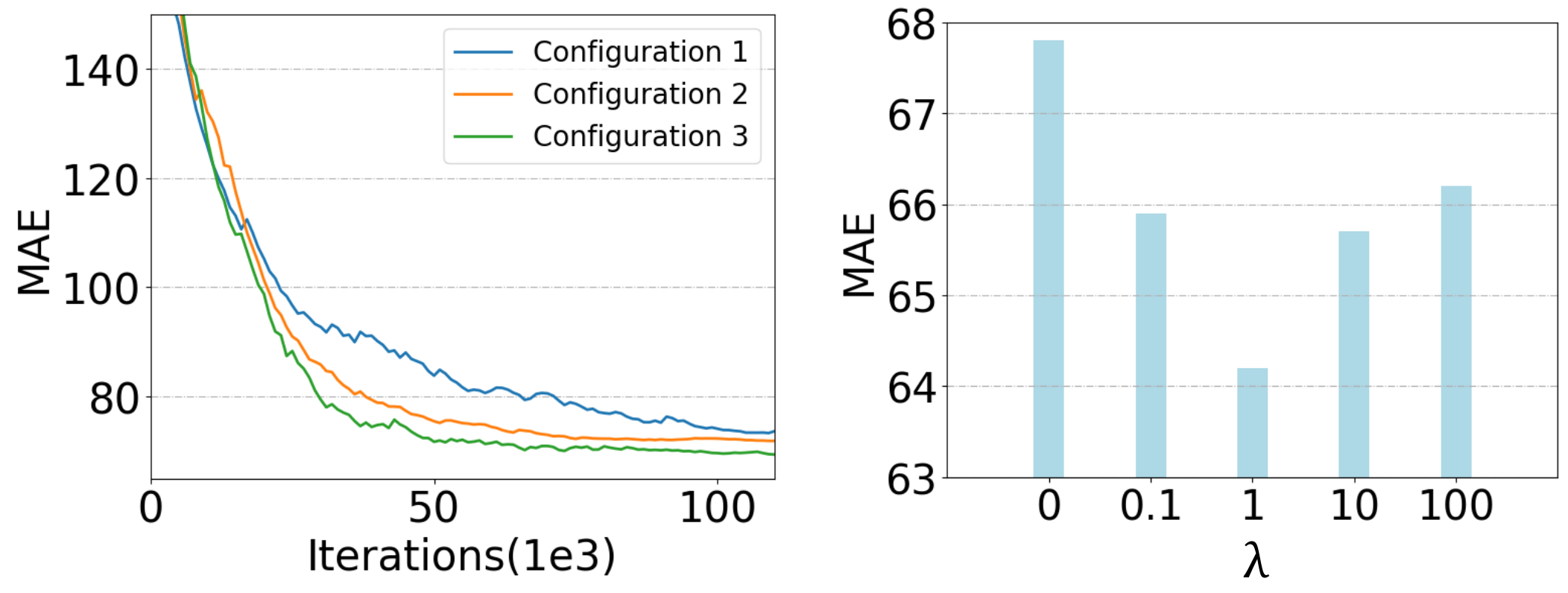}
\caption{Illustration of convergence performance for different network structures and supervision losses.}
\label{fig:AblaPic}
\vspace{-6mm}
\end{figure}
\subsection{Ablation Study}
\label{sec:42}
\vspace{-2mm}
The ablation study results are shown in Table \ref{tab:ablatation}. The table is partitioned row-wise into three groups, with five configurations. Each group contains the indexed configurations corresponding to one main contribution of TEDnet. These include the trellis network with multi-path decoding, the distributed supervision, and the combinatorial loss with SAL and SCL. In different columns, we report the counting accuracy of each configuration, using the MAE metric. We also illustrate the quality of the density map using the PSNR metrics. In Figure \ref{fig:AblaPic}, the left picture illustrates the convergence performance of configurations 1, 2, and 3, demonstrating the benefits for convergence introduced by the dense skip connections and the distributed supervision. The right picture in Figure \ref{fig:AblaPic} shows that when $\lambda = 1$, i.e. the SAL and SCL are equally weighted in the combinatorial loss, the best counting accuracy is reported.

The first group of configurations shown in Table \ref{tab:ablatation} compare the performance of the multi-path trellis decoder and single-path hourglass decoder. Two configurations in this group adopt the same trellis encoder, and single supervision is applied on density map $Z$ with the normal MSE loss. The results in this group show that the multi-path decoder improves the counting accuracy by 2.6\% in terms of MAE metric, and enhance the density map quality by 7.1\% in PSNR metric. Furthermore, the yellow curve in Figure \ref{fig:AblaPic} demonstrates faster convergence thanks to the dense skip connections implemented in the multi-path decoder. The second group of configurations are all set up with TEDnet, using the normal MSE loss. The results show that distributed supervision improves the MAE by 2.8\% and PSNR by 1.9\%, and the green curve shows further improved convergence speed and performance. In the last group, we compare the performance of distributedly deploying different losses. The combinatorial loss with both SAL and SCL ($\lambda=1$) stands out, with 7.2\% improvements in MAE and 4.5\% in PSNR, which confirms that higher density map quality can improve the counting accuracy. Overall, the best result is reported by configuration 5, which incorporates all three contributions.
\vspace{-2mm}
\begin{table}
\caption{Ablation study results on ShanghaiTech Part\underline{ }A dataset. Best performance is bolded. Arrows in all tables indicate the favorable directions of the metric values.}
\vspace{-1mm}
\begin{center}
%\small
\footnotesize
\begin{tabular}{|c|c|c|c|c|}
\hline
\multicolumn{3}{|c|}{Configurations} & MAE$\downarrow$ & PSNR$\uparrow$ \\
\hline
\multirow{4}{*}{\makecell[tl]{Network \\ Structure}}
&\multirow{2}{*}{1} & Trellis Encoder + & \multirow{2}{*}{73.1} & \multirow{2}{*}{22.51} \\
& &Single path Decoder & & \\
\cline{2-5}
&\multirow{2}{*}{2}&Trellis Encoder + Multi- & \multirow{2}{*}{71.2} & \multirow{2}{*}{24.24} \\
& &path Trellis Decoder & & \\ \hline
\multirow{4}{*}{\makecell[tl]{Supervision \\  Methodology}}
&\multirow{2}{*}{2}& Single & \multirow{2}{*}{71.2} & \multirow{2}{*}{24.24} \\
& & Supervision & & \\ %\hline
\cline{2-5}
&\multirow{2}{*}{3}& Distributed & \multirow{2}{*}{69.2} & \multirow{2}{*}{24.71} \\ %\hline
& & Supervision & & \\ \hline
\multirow{3}{*}{Loss Function}
&3& Normal MSE & 69.2 & 24.71\\
\cline{2-5}
&4& SAL & 67.8 & 24.94 \\
\cline{2-5}
&5& SAL + SCL & \textbf{64.2} & \textbf{25.88} \\ \hline
\end{tabular}
\end{center}
\label{tab:ablatation}
\vspace{-8mm}
\end{table}
\subsection{Performance and Comparison}
\vspace{-2mm}
\begin{table*}
\caption{Estimation errors on the ShanghaiTech dataset, the UCF\underline{ }CC\underline{ }50 and the UCF-QNRF dataset}
\vspace{1mm}
\begin{center}
%\small
\footnotesize
\begin{tabular}{|l|cc|cc|cc|cc|}
\hline
\qquad & \multicolumn{2}{|c|}{ShanghaiTech Part\underline{ }A} & \multicolumn{2}{|c|}{ShanghaiTech Part\underline{ }B} & \multicolumn{2}{|c|}{UCF\underline{ }CC\underline{ }50} & \multicolumn{2}{|c|}{UCF-QNRF}\\ \hline
Method & MAE$\downarrow$ & MSE$\downarrow$ & MAE$\downarrow$ & MSE$\downarrow$ & MAE$\downarrow$ & MSE$\downarrow$ & MAE$\downarrow$ & MSE$\downarrow$\\
\hline\hline
Zhang et al.\cite{CNNpatch} & 181.8 & 277.7 & 32.0 & 49.8 & 467.0 & 498.5 & - & - \\
MCNN \cite{MCNN}& 110.2 & 173.2 & 26.4 & 41.3 & 377.6 & 509.1 & 277 & 426\\ %\hline
Cascaded-MTL \cite{sindagi2017cnn}& 101.3 & 152.4 & 20.0 & 31.1 & 322.8 & 397.9 & 252 & 514\\
Switching-CNN \cite{SwitchingCNN} & 90.4 & 135.0 & 21.6 & 33.4 & 318.1 & 439.2 & 228 & 445\\
CP-CNN \cite{PyramidCNNsICCV2017}& 73.6 & 106.4 & 20.1 & 30.1 & 295.8 & \textbf{320.9} & - & -\\ %\hline
CSRNet \cite{CVPR2018CSRNet}& 68.2 & 115.0 & 10.6 & 16.0 & 266.1 & 397.5 & - & -\\ %\hline
SANet \cite{ECCV2018SANet}& 67.0 & \textbf{104.5} & 8.4 & 13.6 & 258.4 & 334.9 & - & -\\ %\hline
Idrees et al. \cite{Arxiv2018CompositionLoss}& - & - & - & - & - & - & 132 & 191\\ %\hline
Ours & \textbf{64.2} & 109.1 & \textbf{8.2} & \textbf{12.8} & \textbf{249.4} & 354.5 & \textbf{113} & \textbf{188}\\ \hline
\end{tabular}
\end{center}
\label{tab:2}
\vspace{-6mm}
\end{table*}

We compare the performance of our TEDnet with eight state-of-the-art approaches, on four challenging datasets, including the ShanghaiTech \cite{MCNN}, the UCF\underline{ }CC\underline{ }50 \cite{Shah2013RCC}, WorldExpo'10 \cite{CNNpatch} and the UCF-QNRF \cite{Arxiv2018CompositionLoss}. We explain the superior performance of TEDnet in terms of both counting accuracy (MAE and MSE as shown in Table \ref{tab:2} and \ref{tab:3}) and density map quality measures (SSIM and PSNR as shown in Table \ref{tab:SSIM}).

\vspace{-4mm}
\subsubsection{Counting Accuracy}
\vspace{-2mm}
\textbf{ShanghaiTech.} The ShanghaiTech dataset is one of the largest datasets which includes Part\underline{ }A and Part\underline{ }B subsets. As shown in Table \ref{tab:2}, on Part\underline{ }A, our method achieves the lowest MAE and a competitive MSE. In terms of MAE, we lead the second best by 4.2\%. On Part\underline{ }B, we report the best performance in terms of both two metrics, where MSE is improved by 5.9\%. The significant improvements on this dataset validate the effectiveness of TEDnet. %which can provide high-quality density maps (as shown in Table \ref{tab:ablatation}) and consequently improve the crowd counting accuracy.
\vspace{1mm}

\noindent\textbf{UCF\underline{ }CC\underline{ }50.} The UCF\underline{ }CC\underline{ }50 dataset introduced by Idrees et al.\cite{Shah2013RCC} contains 50 images of varying resolutions, with a wide range of densities. To settle the sample scarcity problem, we perform a 5-fold cross-validation, following the standard setting in \cite{Shah2013RCC}. As shown in Table \ref{tab:2}, we achieve an improvement of 3.5\% in terms of the MAE metric.
\vspace{1mm}

\begin{table}
\caption{The MAE of the WorldExpo'10 dataset, S is short for Scene.}
\vspace{-2mm}
\begin{center}
%\small
\footnotesize
\begin{tabular}{|l|c|c|c|c|c|c|}
\hline
Method & S1 & S2 & S3 & S4  & S5 & Ave. \\
\hline\hline
%Chen et al. & 2.1 & 55.9 & \textbf{9.6} & 11.3 & 3.4 & 16.5\\
Zhang et al. \cite{CNNpatch} & 9.8 & 14.1 & 14.3 & 22.2 & 3.7 & 12.9\\
MCNN \cite{MCNN}& 3.4 & 20.6 & 12.9 & 13.0 & 8.1 & 11.6\\ %\hline
Switching-CNN \cite{SwitchingCNN}& 4.4 & 15.7 & 10.0 & 11.0 & 5.9 & 9.4\\
CP-CNN \cite{PyramidCNNsICCV2017}& 2.9 & 14.7 & 10.5 & \textbf{10.4} & 5.8 & 8.9\\
CRSNet \cite{CVPR2018CSRNet}& 2.9 & 11.5 & \textbf{8.6} & 16.6 & 3.4 & 8.6\\
SANet \cite{ECCV2018SANet}& 2.6 & 13.2 & 9.0 & 13.3 & 3.0 & 8.2\\
Ours & \textbf{2.3} & \textbf{10.1} & 11.3 & 13.8 & \textbf{2.6} & \textbf{8.0}\\ \hline
\end{tabular}
\end{center}
\label{tab:3}
\vspace{-8mm}
\end{table}

\noindent\textbf{UCF-QNRF.} The UCF-QNRF is a new dataset with one of the highest number of high-count crowd images and annotations. We compare our result with four state-of-the-art methods and our method achieves the best performance in terms of both MAE and MSE. We beat the second best approach by a 14.4\% improvement in MAE and 1.6\% improvement in MSE, as shown in Table \ref{tab:2}.
\vspace{1mm}

\noindent\textbf{WorldExpo¡¯10.} The WorldExpo¡¯10 dataset was introduced by Zhang et al. \cite{CNNpatch}, containing 3980 frames from 108 different scenes from the Shanghai 2010 WorldExpo. Table~\ref{tab:3} shows that TEDnet delivers the lowest MAE in 3 out of 5 test scenes, and reports up to 13.3\% improvement in Scene 5 over others. Overall, we achieve the best average MAE performance, outperforming the second best by 2.4\%.
\vspace{-4mm}

\subsubsection{Density Map Quality}
\vspace{-2mm}
As mentioned in Section \ref{sec:related}, CP-CNN \cite{PyramidCNNsICCV2017} and CSRnet \cite{CVPR2018CSRNet} also emphasize generating high-quality density maps. MCNN \cite{MCNN} is one of the most representative methods in density estimation based crowd counting. We compare the quality of density maps estimated by TEDnet and these three state-of-the-art systems. Quantitatively, as demonstrated in Table \ref{tab:SSIM}, our method outperforms the other methods in both PSNR and SSIM metrics on the ShanghaiTech Part\underline{ }A dataset. Particularly, we obtain 8.1\% and 8.4\% improvements over the second best method, in terms of PSNR and SSIM metrics. Qualitatively, we visualize the maps generated by MCNN, SANet, and TEDnet on the ShanghaiTech Part\underline{ }A in Figure \ref{fig:demo}. In addition, we also display the density maps generated by TEDnet on other datasets in Figure \ref{fig:LastDemo}.
\begin{table}
\caption{Quality of density map on ShanghaiTech Part\underline{ }A dataset and parameter studies, M stands for millions.}
\vspace{-2mm}
\begin{center}
\footnotesize
\begin{tabular}{|l|c|c|c|}
\hline
Method & PSNR$\uparrow$ & SSIM$\uparrow$ & Parameters\\
\hline
MCNN \cite{MCNN} & 21.4 & 0.52 & 0.13M\\ %\hline
CP-CNN \cite{PyramidCNNsICCV2017}& 21.72 & 0.72 & 68.4M\\
CRSNet \cite{CVPR2018CSRNet}& 23.79 & 0.76 & 16.26M\\
Ours & \textbf{25.88} & \textbf{0.83} & \textbf{1.63M}\\ \hline
\end{tabular}
\end{center}
\label{tab:SSIM}
\vspace{-8mm}
\end{table}

Our TEDnet introduces an enhanced multi-path decoder architecture, which, however, is still lightweight compared to other state-of-the-art methods, which also strives to generate high-quality density maps. As shown in Table \ref{tab:SSIM}, the number of parameters in TEDnet is only equal to 10\% of those in CRSNet and 2.4\% of CP-CNN. More importantly, we demonstrate the best overall performance in density map quality as well as counting accuracy. MCNN is most lightweight, yet we show significant improvement in PSNR by 17.3\% and 36\% in SSIM. Moreover, we also outperform MCNN on all datasets in terms of MAE and MSE.
\vspace{-3mm}

\section{Conclusion}
\vspace{-2mm}
In this paper, we have presented a new deep learning architecture, called the trellis encoder-decoder network (TEDnet) for crowd counting. It consists of a multi-scale encoder and a multi-path decoder to generate high-quality density estimation maps. It preserves the localization precision in the encoded feature maps, upon which a multi-path decoder with dense skip connections is adopted to achieve thorough aggregation and fusion of multi-scale features. The TEDnet is trained with the distributed supervision implemented with the proposed combinatorial loss. Experiments on four benchmarks show that the TEDnet achieves new state-of-the-art performance in terms of both density map quality and crowd counting accuracy.
\vspace{-4mm}
\section{Acknowledgment}
\vspace{-2mm}
This paper was supported by the National Science Fund for Distinguished Young Scholars under Grant 61425014, the National Key Scientific Instrument and Equipment Development Project under Grant 61827901, and the Natural Science Foundation of China under Grant 91538204, 61871016.
%-------------------------------------------------------------------------

{\small
\bibliographystyle{ieee}
\bibliography{egbib}

\begin{thebibliography}{10}\itemsep=-1pt

\bibitem{Crowdnet}
L.~Boominathan, S.~S. Kruthiventi, and R.~V. Babu.
\newblock Crowdnet: A deep convolutional network for dense crowd counting.
\newblock In {\em Proceedings of the 2016 ACM on Multimedia Conference}, pages
  640--644. ACM, 2016.

\bibitem{ECCV2018SANet}
X.~Cao, Z.~Wang, Y.~Zhao, and F.~Su.
\newblock Scale aggregation network for accurate and efficient crowd counting.
\newblock In {\em Proceedings of the European Conference on Computer Vision
  (ECCV)}, pages 734--750, 2018.

\bibitem{RCC1}
A.~B. Chan and N.~Vasconcelos.
\newblock Bayesian poisson regression for crowd counting.
\newblock In {\em Computer Vision, 2009 IEEE 12th International Conference on},
  pages 545--551. IEEE, 2009.

\bibitem{RidgeRegressionCC}
K.~Chen, C.~C. Loy, S.~Gong, and T.~Xiang.
\newblock Feature mining for localised crowd counting.
\newblock In {\em BMVC}, volume~1, page~3, 2012.

\bibitem{2012HumanDetection}
P.~Dollar, C.~Wojek, B.~Schiele, and P.~Perona.
\newblock Pedestrian detection: An evaluation of the state of the art.
\newblock {\em IEEE transactions on pattern analysis and machine intelligence},
  34(4):743--761, 2012.

\bibitem{SSFSRCNN}
C.~Dong, C.~C. Loy, and X.~Tang.
\newblock Accelerating the super-resolution convolutional neural network.
\newblock In {\em European Conference on Computer Vision}, pages 391--407.
  Springer, 2016.

\bibitem{DensityCCNoDL2}
L.~Fiaschi, U.~K{\"o}the, R.~Nair, and F.~A. Hamprecht.
\newblock Learning to count with regression forest and structured labels.
\newblock In {\em Pattern Recognition (ICPR), 2012 21st International
  Conference on}, pages 2685--2688. IEEE, 2012.

\bibitem{SDN}
J.~Fu, J.~Liu, Y.~Wang, and H.~Lu.
\newblock Stacked deconvolutional network for semantic segmentation.
\newblock {\em arXiv preprint arXiv:1708.04943}, 2017.

\bibitem{2009DCC}
W.~Ge and R.~T. Collins.
\newblock Marked point processes for crowd counting.
\newblock In {\em Computer Vision and Pattern Recognition, 2009. CVPR 2009.
  IEEE Conference on}, pages 2913--2920. IEEE, 2009.

\bibitem{Shah2013RCC}
H.~Idrees, I.~Saleemi, C.~Seibert, and M.~Shah.
\newblock Multi-source multi-scale counting in extremely dense crowd images.
\newblock In {\em Proceedings of the IEEE conference on computer vision and
  pattern recognition}, pages 2547--2554, 2013.

\bibitem{Arxiv2018CompositionLoss}
H.~Idrees, M.~Tayyab, K.~Athrey, D.~Zhang, S.~Al-Maadeed, N.~Rajpoot, and
  M.~Shah.
\newblock Composition loss for counting, density map estimation and
  localization in dense crowds.
\newblock {\em arXiv preprint arXiv:1808.01050}, 2018.

\bibitem{wacvTrack}
X.~Jiang, P.~Li, X.~Zhen, and X.~Cao.
\newblock Model-free tracking with deep appearance and motion features
  integration.
\newblock In {\em 2019 IEEE Winter Conference on Applications of Computer
  Vision (WACV)}, pages 101--110. IEEE, 2019.

\bibitem{beyondcounting}
D.~Kang, Z.~Ma, and A.~B. Chan.
\newblock Beyond counting: comparisons of density maps for crowd analysis
  tasks-counting, detection, and tracking.
\newblock {\em IEEE Transactions on Circuits and Systems for Video Technology},
  2018.

\bibitem{GTGen}
D.~P. Kingma and J.~Ba.
\newblock Adam: A method for stochastic optimization.
\newblock {\em arXiv preprint arXiv:1412.6980}, 2014.

\bibitem{MoCNN}
S.~Kumagai, K.~Hotta, and T.~Kurita.
\newblock Mixture of counting cnns: Adaptive integration of cnns specialized to
  specific appearance for crowd counting.
\newblock {\em arXiv preprint arXiv:1703.09393}, 2017.

\bibitem{MultiSup2a}
C.-Y. Lee, S.~Xie, P.~Gallagher, Z.~Zhang, and Z.~Tu.
\newblock Deeply-supervised nets.
\newblock In {\em Artificial Intelligence and Statistics}, pages 562--570,
  2015.

\bibitem{FirstDmapCC}
V.~Lempitsky and A.~Zisserman.
\newblock Learning to count objects in images.
\newblock In {\em Advances in neural information processing systems}, pages
  1324--1332, 2010.

\bibitem{2008DCC}
M.~Li, Z.~Zhang, K.~Huang, and T.~Tan.
\newblock Estimating the number of people in crowded scenes by mid based
  foreground segmentation and head-shoulder detection.
\newblock In {\em Pattern Recognition, 2008. ICPR 2008. 19th International
  Conference on}, pages 1--4. IEEE, 2008.

\bibitem{CVPR2018CSRNet}
Y.~Li, X.~Zhang, and D.~Chen.
\newblock Csrnet: Dilated convolutional neural networks for understanding the
  highly congested scenes.
\newblock In {\em Proceedings of the IEEE Conference on Computer Vision and
  Pattern Recognition}, pages 1091--1100, 2018.

\bibitem{Refinenet}
G.~Lin, A.~Milan, C.~Shen, and I.~D. Reid.
\newblock Refinenet: Multi-path refinement networks for high-resolution
  semantic segmentation.
\newblock In {\em Cvpr}, volume~1, page~5, 2017.

\bibitem{Decidenet}
J.~Liu, C.~Gao, D.~Meng, and A.~G. Hauptmann.
\newblock Decidenet: Counting varying density crowds through attention guided
  detection and density estimation.
\newblock In {\em Proceedings of the IEEE Conference on Computer Vision and
  Pattern Recognition}, pages 5197--5206, 2018.

\bibitem{Arxiv2018LeveragingUnlabeld}
X.~Liu, J.~van~de Weijer, and A.~D. Bagdanov.
\newblock Leveraging unlabeled data for crowd counting by learning to rank.
\newblock {\em arXiv preprint arXiv:1803.03095}, 2018.

\bibitem{LaoxieACCV}
E.~Lu, W.~Xie, and A.~Zisserman.
\newblock Class-agnostic counting.
\newblock {\em arXiv preprint arXiv:1811.00472}, 2018.

\bibitem{ccdetect2015cvpr}
Z.~Ma, L.~Yu, and A.~B. Chan.
\newblock Small instance detection by integer programming on object density
  maps.
\newblock In {\em Proceedings of the IEEE Conference on Computer Vision and
  Pattern Recognition}, pages 3689--3697, 2015.

\bibitem{Refinenet2}
V.~Nekrasov, C.~Shen, and I.~Reid.
\newblock Light-weight refinenet for real-time semantic segmentation.
\newblock {\em arXiv preprint arXiv:1810.03272}, 2018.

\bibitem{SISRupample}
A.~Odena, V.~Dumoulin, and C.~Olah.
\newblock Deconvolution and checkerboard artifacts.
\newblock {\em Distill}, 2016.

\bibitem{HydraCNN}
D.~Onoro-Rubio and R.~J. L{\'o}pez-Sastre.
\newblock Towards perspective-free object counting with deep learning.
\newblock In {\em European Conference on Computer Vision}, pages 615--629.
  Springer, 2016.

\bibitem{LinearRegressionCC}
N.~Paragios and V.~Ramesh.
\newblock A mrf-based approach for real-time subway monitoring.
\newblock In {\em Computer Vision and Pattern Recognition, 2001. CVPR 2001.
  Proceedings of the 2001 IEEE Computer Society Conference on}, volume~1, pages
  I--I. IEEE, 2001.

\bibitem{DensityCCNoDL3}
V.-Q. Pham, T.~Kozakaya, O.~Yamaguchi, and R.~Okada.
\newblock Count forest: Co-voting uncertain number of targets using random
  forest for crowd density estimation.
\newblock In {\em Proceedings of the IEEE International Conference on Computer
  Vision}, pages 3253--3261, 2015.

\bibitem{cctracking}
W.~Ren, D.~Kang, Y.~Tang, and A.~B. Chan.
\newblock Fusing crowd density maps and visual object trackers for people
  tracking in crowd scenes.
\newblock In {\em Proceedings of the IEEE Conference on Computer Vision and
  Pattern Recognition}, pages 5353--5362, 2018.

\bibitem{DTCC}
M.~Rodriguez, I.~Laptev, J.~Sivic, and J.-Y. Audibert.
\newblock Density-aware person detection and tracking in crowds.
\newblock In {\em Computer Vision (ICCV), 2011 IEEE International Conference
  on}, pages 2423--2430. IEEE, 2011.

\bibitem{cctracking2011iccv}
M.~Rodriguez, I.~Laptev, J.~Sivic, and J.-Y. Audibert.
\newblock Density-aware person detection and tracking in crowds.
\newblock In {\em Computer Vision (ICCV), 2011 IEEE International Conference
  on}, pages 2423--2430. IEEE, 2011.

\bibitem{SegUnet}
O.~Ronneberger, P.~Fischer, and T.~Brox.
\newblock U-net: Convolutional networks for biomedical image segmentation.
\newblock In {\em International Conference on Medical image computing and
  computer-assisted intervention}, pages 234--241. Springer, 2015.

\bibitem{RCC2}
D.~Ryan, S.~Denman, C.~Fookes, and S.~Sridharan.
\newblock Crowd counting using multiple local features.
\newblock In {\em Digital Image Computing: Techniques and Applications, 2009.
  DICTA'09.}, pages 81--88. IEEE, 2009.

\bibitem{Survey2015a}
D.~Ryan, S.~Denman, S.~Sridharan, and C.~Fookes.
\newblock An evaluation of crowd counting methods, features and regression
  models.
\newblock {\em Computer Vision and Image Understanding}, 130:1--17, 2015.

\bibitem{Survey2015b}
S.~A.~M. Saleh, S.~A. Suandi, and H.~Ibrahim.
\newblock Recent survey on crowd density estimation and counting for visual
  surveillance.
\newblock {\em Engineering Applications of Artificial Intelligence},
  41:103--114, 2015.

\bibitem{SwitchingCNN}
D.~B. Sam, S.~Surya, and R.~V. Babu.
\newblock Switching convolutional neural network for crowd counting.
\newblock In {\em Proceedings of the IEEE Conference on Computer Vision and
  Pattern Recognition}, volume~1, page~6, 2017.

\bibitem{CVPR18DeepNegativeCorrelation}
Z.~Shi, L.~Zhang, Y.~Liu, X.~Cao, Y.~Ye, M.-M. Cheng, and G.~Zheng.
\newblock Crowd counting with deep negative correlation learning.
\newblock In {\em Proceedings of the IEEE Conference on Computer Vision and
  Pattern Recognition}, pages 5382--5390, 2018.

\bibitem{sindagi2017cnn}
V.~A. Sindagi and V.~M. Patel.
\newblock Cnn-based cascaded multi-task learning of high-level prior and
  density estimation for crowd counting.
\newblock In {\em Advanced Video and Signal Based Surveillance (AVSS), 2017
  14th IEEE International Conference on}, pages 1--6. IEEE, 2017.

\bibitem{PyramidCNNsICCV2017}
V.~A. Sindagi and V.~M. Patel.
\newblock Generating high-quality crowd density maps using contextual pyramid
  cnns.
\newblock In {\em 2017 IEEE International Conference on Computer Vision
  (ICCV)}, pages 1879--1888. IEEE, 2017.

\bibitem{Survey2018}
V.~A. Sindagi and V.~M. Patel.
\newblock A survey of recent advances in cnn-based single image crowd counting
  and density estimation.
\newblock {\em Pattern Recognition Letters}, 107:3--16, 2018.

\bibitem{SS2}
X.~Tao, H.~Gao, R.~Liao, J.~Wang, and J.~Jia.
\newblock Detail-revealing deep video super-resolution.
\newblock In {\em Proceedings of the IEEE International Conference on Computer
  Vision, Venice, Italy}, pages 22--29, 2017.

\bibitem{CFNet}
J.~Valmadre, L.~Bertinetto, J.~Henriques, A.~Vedaldi, and P.~H. Torr.
\newblock End-to-end representation learning for correlation filter based
  tracking.
\newblock In {\em Computer Vision and Pattern Recognition (CVPR), 2017 IEEE
  Conference on}, pages 5000--5008. IEEE, 2017.

\bibitem{CNNboost}
E.~Walach and L.~Wolf.
\newblock Learning to count with cnn boosting.
\newblock In {\em European Conference on Computer Vision}, pages 660--676.
  Springer, 2016.

\bibitem{DCC2}
M.~Wang and X.~Wang.
\newblock Automatic adaptation of a generic pedestrian detector to a specific
  traffic scene.
\newblock In {\em Computer Vision and Pattern Recognition (CVPR), 2011 IEEE
  Conference on}, pages 3401--3408. IEEE, 2011.

\bibitem{ssim}
Z.~Wang, A.~C. Bovik, H.~R. Sheikh, and E.~P. Simoncelli.
\newblock Image quality assessment: from error visibility to structural
  similarity.
\newblock {\em IEEE transactions on image processing}, 13(4):600--612, 2004.

\bibitem{ourcc}
Z.~Wang, Z.~Xiao, K.~Xie, Q.~Qiu, X.~Zhen, and X.~Cao.
\newblock In defense of single-column networks for crowd counting.
\newblock {\em arXiv preprint arXiv:1808.06133}, 2018.

\bibitem{MultiSup1hed}
S.~Xie and Z.~Tu.
\newblock Holistically-nested edge detection.
\newblock In {\em Proceedings of the IEEE international conference on computer
  vision}, pages 1395--1403, 2015.

\bibitem{DensityCCNoDL4}
B.~Xu and G.~Qiu.
\newblock Crowd density estimation based on rich features and random projection
  forest.
\newblock In {\em Applications of Computer Vision (WACV), 2016 IEEE Winter
  Conference on}, pages 1--8. IEEE, 2016.

\bibitem{SDN2}
J.~Yang, Q.~Liu, and K.~Zhang.
\newblock Stacked hourglass network for robust facial landmark localisation.
\newblock In {\em Computer Vision and Pattern Recognition Workshops (CVPRW),
  2017 IEEE Conference on}, pages 2025--2033. IEEE, 2017.

\bibitem{ZhangCNNPatch}
C.~Zhang, H.~Li, X.~Wang, and X.~Yang.
\newblock Cross-scene crowd counting via deep convolutional neural networks.
\newblock In {\em Proceedings of the IEEE Conference on Computer Vision and
  Pattern Recognition}, pages 833--841, 2015.

\bibitem{CNNpatch}
C.~Zhang, H.~Li, X.~Wang, and X.~Yang.
\newblock Cross-scene crowd counting via deep convolutional neural networks.
\newblock In {\em Proceedings of the IEEE Conference on Computer Vision and
  Pattern Recognition}, pages 833--841, 2015.

\bibitem{MCNN}
Y.~Zhang, D.~Zhou, S.~Chen, S.~Gao, and Y.~Ma.
\newblock Single-image crowd counting via multi-column convolutional neural
  network.
\newblock In {\em Proceedings of the IEEE conference on computer vision and
  pattern recognition}, pages 589--597, 2016.

\bibitem{SegPSPnet}
H.~Zhao, J.~Shi, X.~Qi, X.~Wang, and J.~Jia.
\newblock Pyramid scene parsing network.
\newblock In {\em IEEE Conf. on Computer Vision and Pattern Recognition
  (CVPR)}, pages 2881--2890, 2017.

\end{thebibliography}
}

\end{document}